# An Investigation of Traffic Density Changes inside Wuhan during the COVID-19 Epidemic with GF-2 Time-Series Images

Chen Wu[1], Yinong Guo[1], Haonan Guo[1], Jingwen Yuan[2], Lixiang Ru[3], Hongruixuan Chen[1], Bo Du[3], Liangpei Zhang[1]

1. State Key Laboratory of Information Engineering in Surveying, Mapping and Remote Sensing, Wuhan University

2. School of Remote Sensing and Information Engineering, Wuhan University

3. School of Computer Science, Wuhan University

Corresponding author: Bo Du, Liangpei Zhang

E-mail: gunspace@163.com, zlp62@whu.edu.cn

## Abstract

In order to mitigate the spread of COVID-19, Wuhan was the first city to implement strict lockdown policy in 2020. Even though numerous researches have discussed the travel restriction between cities and provinces, few studies focus on the effect of transportation control inside the city due to the lack of the measurement and available data in Wuhan. Since the public transports have been shut down in the beginning of city lockdown, the change of traffic density is a good indicator to reflect the intracity population flow. Therefore, in this paper, we collected time-series high-resolution remote sensing images with the resolution of 1m acquired before, during and after

Wuhan lockdown by GF-2 satellite. Vehicles on the road were extracted and counted for the statistics of traffic density to reflect the changes of human transmissions in the whole period of Wuhan lockdown. Open Street Map was used to obtain observation road surfaces, and a vehicle detection method combing morphology filter and deep learning was utilized to extract vehicles with the accuracy of 62.56%. According to the experimental results, the traffic density of Wuhan dropped with the percentage higher than 80%, and even higher than 90% on main roads during city lockdown; after lockdown lift, the traffic density recovered to the normal rate. Traffic density distributions also show the obvious reduction and increase throughout the whole study area. The significant reduction and recovery of traffic density indicates that the lockdown policy in Wuhan show effectiveness in controlling human transmission inside the city, and the city returned to normal after lockdown lift.

## 1. Introduction

In January 2020, the novel coronavirus disease 2019 (COVID-19) caused by SARS-CoV-2 quickly spread in the city of Wuhan, China (China Central Television 2019). Since there is no specific drug treatment and vaccine, the government implemented the travel ban in Wuhan city to control the epidemic (Chen et al. 2020; Tian et al. 2020). Wuhan shut down all the inbound and outbound transportations, as well as the public transports inside the city, on 23 January, 2020 (Wuhan municipal headquarters for the COVID-19 epidemic prevention and control 2020b). It appears to be the largest quarantine in human history, considering the policy covered more than 9 million people in Wuhan before city lockdown (China News 2020; Tian et al. 2020).

The lockdown policies in Wuhan became stricter during the COVID-19 epidemic, where all motor vehicles were banned in central urban area on 26 January (Wuhan municipal headquarters for the COVID-19 epidemic prevention and control 2020c), and all residential communities restricted the access in and out on 10 February (Wuhan municipal headquarters for the COVID-19 epidemic prevention and control 2020a). Until 8 April, Wuhan lifted lockdown after 76 days (Hubei municipal headquarters for the COVID-19 epidemic prevention and control 2020).

In lockdown policy, there are mainly two aspects: travel ban between cities, which can mitigate the spread of COVID-19 to other regions; and transportation control inside the city, which is implemented to control intracity population transmission to slow down the increase of the infections. Lots of researches have indicated that the transmission control measures limited the growth of the COVID-19 epidemic in China (Chen et al. 2020; Chinazzi et al. 2020; Jia et al. 2020; Kraemer et al. 2020; Tian et al. 2020). However, most publications focus on the study of the travel ban from Wuhan to other cities and provinces, which may be because it is hard to quantitatively evaluate the population flow inside the city. Several researches also indicated that human mobility limitation shows correlation with COVID-19 case reduction (Gao et al. 2020), thus it is also valuable to analyze the intracity population flow. Considering the public transports have been shut down, the change of traffic density is a good indicator to reflect the situations of intracity human transmission during the whole period of Wuhan lockdown.

Remote sensing provides an appropriate data source for large-scale study (Huang et al. 2020; Liu et al. 2021; Reichstein et al. 2019), such as environment study (Gray et al. 2020; Liu et al. 2020; Su et al. 2021; Xu et al. 2021), land-use/land-cover study (Gao and O'Neill 2020; Jin et al. 2013; Yang et al. 2018), vegetation monitoring (Axelsson et al. 2021; Mardian et al. 2021; Taubert et al. 2018). Nowadays, the high-resolution remote sensing images show significant potentials to distinguish cars from the road, and make an objective statistic for counting the vehicle number on the road throughout the whole city (Audebert et al. 2017; Chen et al. 2016; Eikvil et al. 2009; Ji et al. 2020; Leitloff et al. 2010; Li et al. 2019; Tang et al. 2017a; Tang et al. 2017b; Tanveer et al. 2020; Tao et al. 2019).

In order to implement an investigation of traffic density changes inside Wuhan caused by COVID-19, we collected high-resolution time-series images acquired during the whole progress of Wuhan lockdown, by a Chinese satellite GF-2 with the resolution as high as 0.8m. These images covered a total area of $1273.61 km^2$. Since the vehicles are nearly a clustering of pixels without any details in such a resolution, we utilize a method combining morphology filtering and deep learning to detect vehicles on the road. The road lines from open street map were used to exclude the non-road interferences. The vehicle numbers before, during and after Wuhan lockdown were counted to measure the effectiveness of the lockdown policy in the whole city, even specific into ring roads and high-level roads. The main contributions of this manuscript can be summarized as follows:

1) A time-series GF-2 image set with the resolution of about 1m was collected covering the whole period of Wuhan lockdown. Vehicles on the roads were detected to extract the traffic densities of Wuhan city before, during and after lockdown.

2) A hybrid model with morphology filter and deep learning identification was utilized to identify vehicles on the road. This method was proved to be effective by considering the shape and background information, dealing with the problem that vehicles show few detailed information with such a low but common resolution.

3) The variations of traffic densities reflect the rapid reduction of population transmission after lockdown implementation with the rate of higher than 90%, and the transportation recovery after lockdown lift.

The manuscript was organized as follows. Section 2 introduces the study site, remote sensing data collected and data pre-processing. In the section 3, we describe the vehicle detection method, including vehicle candidate extraction and deep learning identification. Then, in section 4 and 5, we present detailed analysis about traffic density variation throughout the whole progress of Wuhan lockdown. Finally, we draw a conclusion in section 6.

## 2. Study Site

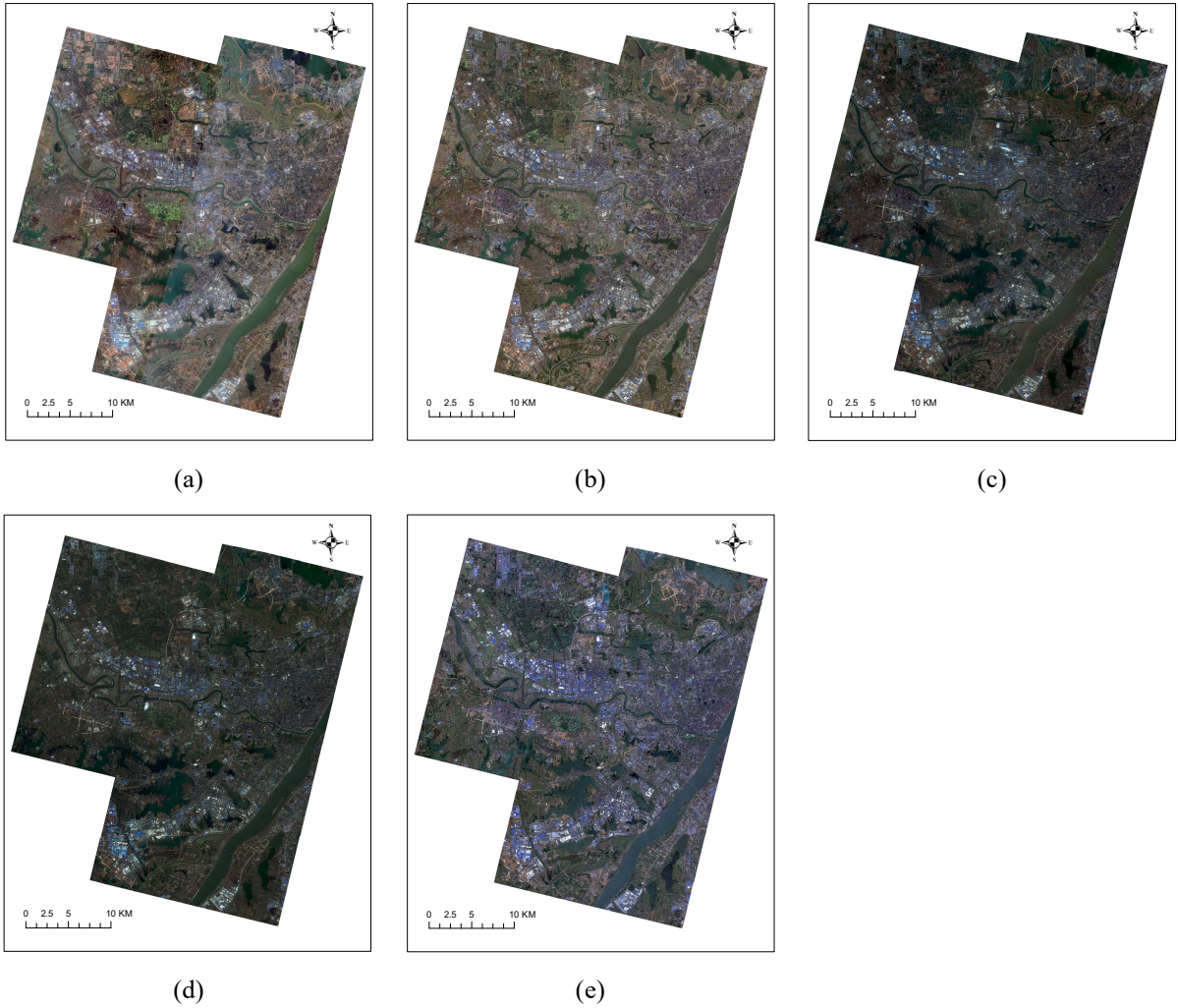

Figure 1. GF-2 high resolution remote sensing images acquired on (a) 28 November, 2018, (b) 19 October, 2019, (c) 30 January, 2020, (d) 9 February, 2020, and (e) 18 May, 2020.

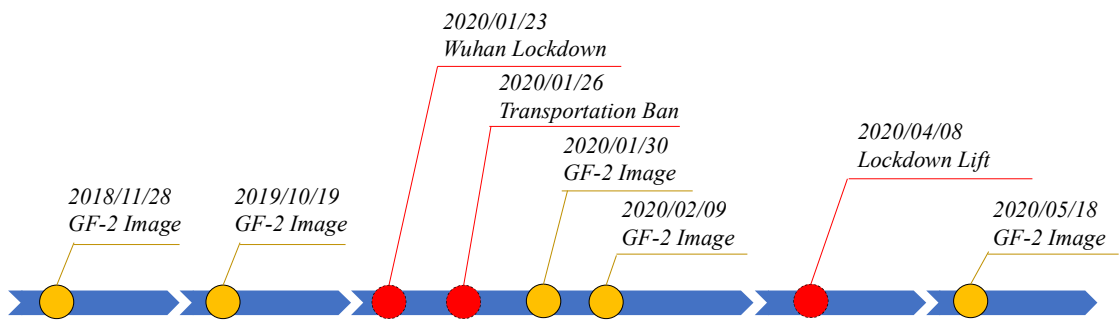

Figure 2. Timeline of Wuhan lockdown and the acquisition time of GF-2 remote sensing images.

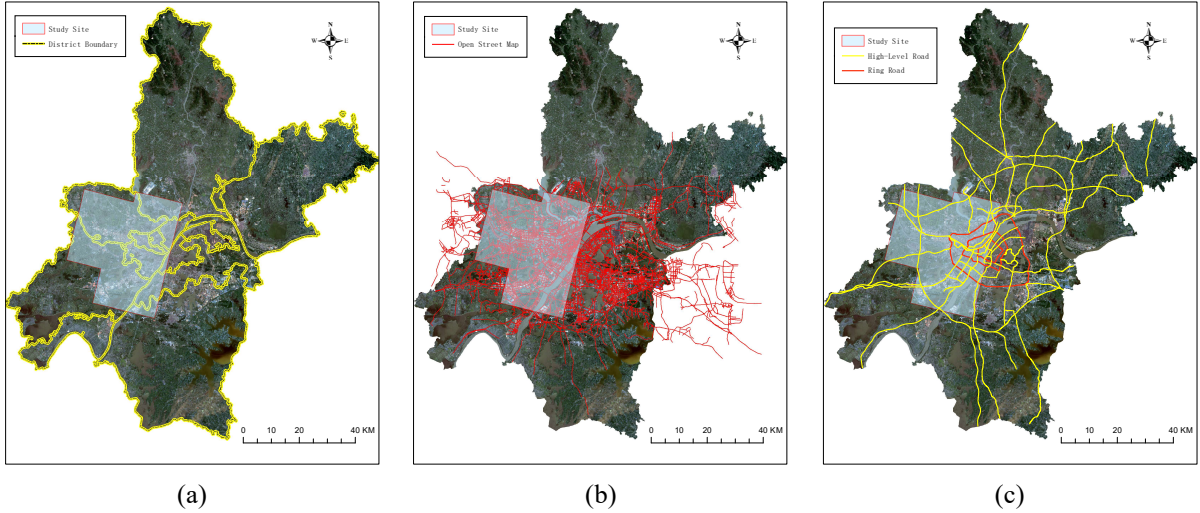

Figure 3. (a) Study site covering Wuhan city; (b) Open street map used for extracting roads; (c) Ring roads and high-level roads for car statistics.

Table 1 Metadata of high-resolution remote sensing images

| Acquisition Date | Week | Image Time | Resolution | Solar Zenith | Satellite Zenith |
|---|---|---|---|---|---|
| 2018/11/28 | Wednesday | 11:41:02 | 0.99m | 52.26° | 80.50° |
| 2019/10/19 | Saturday | 11:28:59 | 0.95m | 41.45° | 83.00° |
| 2020/01/30 | Thursday | 11:17:18 | 0.94m | 51.99° | 63.03° |
| 2020/02/09 | Sunday | 11:19:49 | 1.02m | 49.10° | 69.19° |
| 2020/05/18 | Monday | 11:24:28 | 0.96m | 16.54° | 88.46° |

In order to quantitatively evaluate the traffic density changes during the whole progress of Wuhan lockdown, we collect 5 sets of GF-2 high-resolution remote sensing images, acquired on 28 November, 2018, 19 October, 2019, 30 January, 2020, 9 February, 2020, and 18 May, 2020, separately (as shown in Figure 1). The metadata is shown in Table 1. The acquisition time are before, during, and after Wuhan lockdown, thus these remote sensing images have the ability to reflect the situations throughout the whole progress of COVID-19 epidemic in Wuhan, as shown in Figure 2. The images shown in Figure 1 (a), (c), (e) were all acquired in weekdays, thus are comparable and unaffected to weekends. It is worth noting that since the first 4 images were acquired in winter, the solar zeniths are all large, which means there are large areas of shadows on these images, while the last image has a small solar zenith with few shadow coverages. The different coverages of shadows will result in different observation road areas, which is considered in our study. The satellite zeniths of the two images acquired during lockdown differ from those of the other three images, which lead to more oblique buildings on the images blocking more vehicles on the roads.

Each image set contains several GF-2 high-resolution images, and was mosaic into a large image after carefully co-registration. GF-2 high-resolution image data provide 4 multispectral bands with the resolution of 3.2m, and 1 pan band with the resolution as high as 0.8m. For extracting vehicles on the road, we used the fused images after GS pan-sharpening, and NDVI (Normalized Difference Vegetation Index) images to mask false alarms. The resolutions of pansharpened images are approximately 1m, as listed in Table 1.

The common region of these 5 multi-temporal images was extracted for comparison, as shown in Figure 3 (a). It covers two main parts of Wuhan city (Hankou and Hanyang), and unfortunately, there was few GF-2 image data covering the rest one (Wuchang). The study site has the area of $1273.61 km^2$, which covers 14.79% area of Wuhan city ($1273.61 km^2 / 8608.91 km^2$) and 41.59% area of city center inside the third ring road ($218.06 km^2 / 524.29 km^2$). Therefore, the study site is representative to evaluate the traffic density changes during Wuhan lockdown.

Besides the remote sensing image, we also selected open street map (OSM) of the whole Wuhan city to extract roads, as shown in Figure 3 (b). The five images were all registered to OSM to reduce mis-registration error. The lines with the selected "highway" attributes were used to extract roads, including "motorway", "primary", "secondary", "tertiary", "trunk", "unclassified" and their corresponding "link" (OpenStreetMap Wiki 2020). Buffer with 20m on both sides of OSM road lines were implemented to extract roads from high-resolution images.

Since some roads in city center will be obstructed by high buildings resulting in false alarms, ring roads and high-level roads were also used for a better estimation of traffic density changes, as shown in Figure 3 (c). A buffer with 40m on both sides of ring roads and high-level roads were generated for statistics. These two kinds of roads are categorized into "main roads".

## 3. Methodology for Vehicle Detection

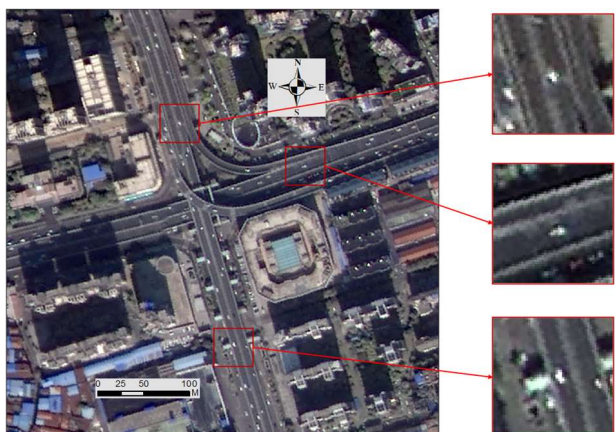

Figure 4. Example of vehicles in GF-2 image with the resolution of 0.9m.

As shown in Figure 4, the vehicles on GF-2 image with the resolution of 0.9m do not show enough shape information for identification. They are very easy to be misclassified with other landscapes. Fortunately, the vehicles on the road have two characteristics: firstly, they are anomalous from the road; secondly, they still contain some characteristics considering their road backgrounds. Therefore, we utilize a vehicle detection method combing candidate extraction by morphology filter and identification by deep learning, which was firstly proposed in our previous work (Wu et al. 2021b).

The flowchart of the vehicle detection method is shown as Figure 5. OSM data was used to separate roads from the image. Morphology filters of top hat and bottom hat were implemented to find bright objects and dark objects. Shape filters with area rule, shape rule and compactness rule were used to filter out unlikely candidates, and anchors were generated according to shape direction and zoom factor. Then, positive samples and negative samples were selected and inputted into the multi-branch deep learning network. The anchors with high possibilities were remained and finally determined with NMS process.

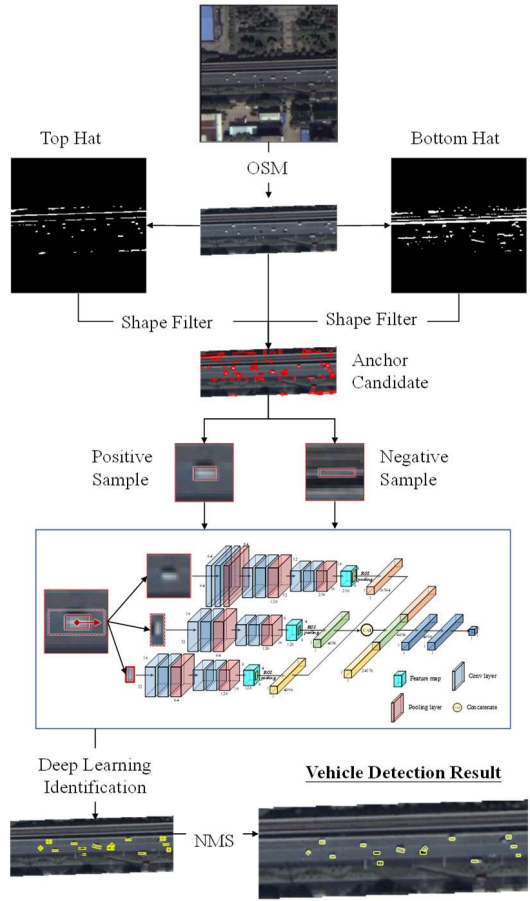

Figure 5.Flowchart of the vehicle detection algorithm.

### 3.1. Vehicle Candidate Extraction

Considering that vehicles on the remote sensing image with the resolution of 0.9m do not contain detailed shape features, it is difficult to detect vehicle objects directly. We decided to utilize the unsupervised method as preprocessing to find vehicle candidates and reduce the interference factors on the road, such as road dividers, pedestrian crossings, highway lane markers, and so on.

We assume that, the vehicles must be significantly different from the roads visually. Therefore, we utilize TopHat and BottomHat morphology process to separate white vehicles and dark vehicles from the roads; then, vehicle shadows adjacent to white vehicles are removed; and finally, these objects are filtered according to their areas, shapes, and compactness. The detailed processes are described as follows:

In this paper, the Open Street Map dataset covering Wuhan city was used to extract roads with a buffer of 20m. For each band, $7 \times 7$ TopHat and BottomHat morphology filters are applied separately to each band of multi-spectral images to separate bright and dark objects from the background. In order to fuse the multispectral information, the TopHat and BottomHat density images are obtained by calculating the $l2-norm$ of the filter results. Two thresholds are determined by visual interpretation to extract binary maps of bright and dark pixels. Besides, NDVI image is calculated and thresholded to obtain vegetation mask for avoiding false alarms.

After masking with NDVI mask, the binary TopHat and BottomHat maps are clustered into bright and dark objects by the connectivity of 8 neighbourhoods. It is worth noting that shadows accompanying bright vehicles will be remained in BottomHat maps and seriously disturb the determination of dark vehicles. Therefore, we remove the dark objects if they are adjacent to bright objects with a assigned proportion such as 30%. At the same time, if the dark objects have a quite good shape alike vehiles, they will be remained for deep learning identification.

Then, the bright and dark objects are filtered by shape filters with some prior rules. There are three kinds of indicators in shape filter: (1) area indicates the area of objects; (2) shape includes the width, length, and aspect ratio indicating the ratio of width and height of the minimum bounding rectangle; and (3) compactness contains filling proportions of contour hull or minimum bounding box.

All the thresholds of shape filter are determined by expert knowledge, and selected to remove the interferences very unlikely to be vehicles. The filter rules are set as follows:

(1) Area rule: The area of objects should be higher than 2 pixels and smaller than 200 pixels;

(2) Shape rule: After the minimum bounding box of objects is obtained, the length should be lower than 28 pixels (25m considering the resolution), the width should be lower than 9 pixels (8m), and the aspect ratio should be lower than 8.

(3) Compactness rule: The filling proportion of contour hull should be higher than 0.9, while the filling proportion of minimum bounding box should be higher than 0.55.

Finally, the bright and dark objects are combined for vehicle candidates. Since these objects may not be very accurate to represent vehicles considering the low resolution of approximately 1m, we generate anchors centering the objects with a set of zoom ratios. The zoom ratio of length for minimum bounding box is [1, 1.5, 2], and that for width is [1, 1.25, 1.5]. The anchors will be also generated in their vertical direction. Only if the aspect ratio is higher than 0.6, the vehicle direction is fixed in one direction. After the anchor generation, each object will contain 9 or 18 anchor candidates, and these anchor candidates are inputted into deep learning network for identification.

*3.2. Deep Learning Identification*

After the vehicle candidates are extracted by morphology filer and shape filter, we utilize a multi-branch CNN model to perform binary classification for refining vehicle detection. since the vehicle doesn't contain enough shape information in such a resolution of 1m and must be identified considering road background, we choose three image patches with different input sizes centered on anchor candidates. The first input is a window patch with the size of $48 \times 48$, which is larger than a vehicle object without doubt and provides background information; the second input is a sub-window patch with the size of $12 \times 24$ aligned with the direction of anchor, which provides the information from the background and the vehicle; the third input is the anchor, which is aligned with the anchor and resized to the same size of $12 \times 24$ for batch process.

The architecture of network derived from vgg16, whereas is simplified due to the small size of input patch. The detailed architecture is shown in Table 2. Since the sizes of sub-window patch and anchor are smaller than the window patch, these two branches contain fewer layers. The three branches are firstly pre-trained with separate fully connected layers, and then these three convolutional branches are connected by concatenating to one new fully connected layer. The output is activated by sigmoid function to be changed (1) or unchanged (0).

Table 2 Deep Learning Network Architecture

| Multi-branch CNN Network Configuration | | |
|---|---|---|
| Input (48 × 48 patch) | Input (12 × 24 patch) | Input (12 × 24 anchor) |
| conv3-64 | conv3-64 | conv3-64 |
| conv3-64 | conv3-64 | conv3-64 |
| maxpool2 | maxpool2 | maxpool2 |
| conv3-128 | conv3-128 | conv3-128 |
| conv3-128 | conv3-128 | conv3-128 |
| maxpool2 | maxpool2 | maxpool2 |
| conv3-256 | | |
| conv3-256 | | |
| maxpool2 | | |
| *Pre-training* | | |
| fc-4096 | fc-2048 | fc-2048 |
| fc-4096 | fc-2048 | fc-2048 |
| fc-1 | fc-1 | fc-1 |
| sigmoid | sigmoid | sigmoid |
| *Training* | | |
| concatenating | | |
| fc-4096 | | |
| fc-4096 | | |
| fc-1 | | |
| sigmoid | | |

All these networks are trained with the same sets of training samples. Since we have extracted lots of vehicle candidates from the images, the vehicle and non-vehicle training samples are all selected from the candidates. The numbers of training samples in each image are shown in Table 3. It can be observed that the numbers of non-vehicle samples in these five images are similar, whereas vehicle samples during Wuhan lockdown are quite fewer. This is because vehicle numbers on the road during lockdown decrease too heavily to find abundant vehicle samples. Random flipping (both vertically and horizontally) is applied with a pre-defined probability as data augmentation.

According to previous research, the model trained with all the samples from multi-temporal images is the most robust (Wu et al. 2021b). The model trained with all samples and fine-tuned to the specific image shows higher detection rate, which may be more suitable for detecting vehicles in the image during Wuhan lockdown. Therefore, in the 5 images, the results are obtained with the model trained with all samples, where the 2 images during Wuhan lockdown are interpreted with the model fine-tuned to the specific image.

Table 3 Training samples for deep learning network in each remote sensing image

| Image Date | Vehicle Samples | Non-Vehicle Samples |
|---|---|---|
| 2018/11/28 | 3272 | 9888 |
| 2019/10/19 | 3200 | 9122 |
| 2020/01/30 | 800 | 11024 |
| 2020/02/09 | 800 | 11010 |
| 2020/05/18 | 3200 | 10045 |
| **Total** | **11272** | **51089** |

Binary cross-entropy is used as the loss function of optimization. In order to balance the sample numbers of vehicles and non-vehicles, a simple weight process is utilized in loss calculation, where $(n_v + n_n)/2n_v$ is assigned to vehicle samples and $(n_v + n_n)/2n_n$ is assigned to non-vehicle samples. In the training procedure, an Adam optimizer (Kingma and Ba 2014) with initial learning rate of $1e^{-4}$ is used. The batch size and number of epochs are set as 200 and 100, respectively. We use a warmup learning rate scheduler strategy (Gotmare et al. 2019). Specifically, the learning rate firstly goes up from $1e^{-4}$ to $1e^{-3}$ linearly in the first 20 epochs and decays by 0.8 every epoch in consequent, while in the optimization of the window patch branch and the whole network, the start learning rate and max learning rate are assigned as $1e^{-5}$ and $1e^{-4}$.

After the training process, the threshold of 0.5 was used in vehicle detection in the whole region to distinguish vehicles and non-vehicles.

### 3.3. Post-processing

After deep learning identification, lots of anchors centering the vehicles will be remained. In order to find the most accurate anchor for each vehicle, we utilize a modified NMS (Non-Maximum Suppression).

In the modified NMS, the anchors with larger width, height and aspect ratio than the shape filter rule will be removed firstly. Then, the Intersection over Union (IOU) and Intersection over Area (IOA) are both used to filter out overlapping anchors. Finally, for the overlapping anchors, the specific anchor, which has the probability smaller than the maximum one with less than 0.05 but shows the minimum area, will be selected as the remaining anchor to better fit the vehicle shape.

In the final detection results, the shadow coverage will lead to an unfair comparison, since the last image after lockdown lift was acquired in summer, and has a quite small solar zenith with fewer shadow coverages than the other 4 images. It means that the observation road area on the last image is larger than those on the first 4 images. Therefore, in order to implement a reasonable analysis, we utilize a method of shadow removal in the statistics of vehicle counts.

In this process, we use Tsai shadow detection method (Tsai 2006) and manually defined thresholds to separate binary shadow coverages in the five remote sensing images. A $3 \times 3$ closing morphology filter and $7 \times 7$ opening morphology filter were implemented on these binary shadow maps to filter out small interferences and retained large shadow regions. The union of all shadow maps was finally used to erase vehicle detection results in all the 5 remote sensing images, to guarantee the same observation road areas for a fair comparison.

## 4. Experimental Results

### *4.1. Accuracy Assessment*

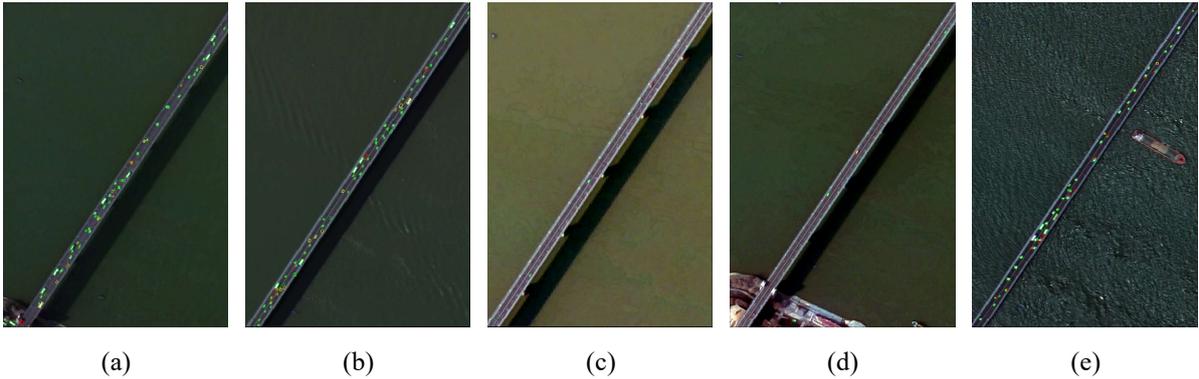

(a)      (b)      (c)      (d)      (e)

Figure 6. Example of accuracy evaluation in a sub-region of the remote sensing images acquired on (a) 28 November, 2018, (b) 19 October, 2019, (c) 30 January, 2020, (d) 9 February, 2020, and (e) 18 May, 2020, where the green rectangles indicate correct detection, the yellow rectangles indicate omission error, and the red rectangles indicate false alarm.

In order to quantitatively evaluate the performance of the proposed vehicle detection algorithm, we selected 9 same regions for all the multi-temporal images with the total coverage area of $75 km^2$, where 5628 vehicles are labelled totally. A small sub-region used in accuracy evaluation is shown in Figure 6, where most vehicles were correctly detected visually, and not many false alarms were remained in the images.

Precision rate, recall rate and F1 score are used for assessment. It can be observed that vehicle candidate extraction firstly extracts most vehicles on the road with a high recall rate, whereas there are many false alarms with a low precision rate. After deep learning identification, most false alarms can be removed with the obviously increased precision rate and slightly reduced recall rate. For the three images before and after lockdown, the accuracies are as high as 70%, while the two images during lockdown only get the accuracy of approximately 47%. The reason is that during Wuhan lockdown, there are only a few vehicles on the road, whereas the interferences causing false alarms still exists. The average accuracy can reach 62% according to accuracy assessment, which is satisfactory considering the resolution of only about 1m.

Table 4 Accuracy Evaluation

|  |  | Vehicle Candidate | Single Image Training | All Images Training without Weighting | All Images Training with Weighting | All Images Training with Single Image Fine-tune |
|---|---|---|---|---|---|---|
| **2018/11/28** | Precision | 9.63% | 54.29% | 73.96% | **77.44%** | 69.31% |
|  | Recall | **89.82%** | 79.87% | 73.53% | 73.88% | 77.37% |
|  | F1 | 17.40% | 64.64% | 73.75% | **75.62%** | 73.12% |
| **2019/10/19** | Precision | 8.63% | **73.45%** | 65.60% | 69.11% | 71.19% |
|  | Recall | **93.61%** | 75.49% | 80.66% | 79.75% | 77.79% |
|  | F1 | 15.81% | 74.45% | 72.35% | 74.05% | **74.34%** |
| **2020/01/30** | Precision | 0.73% | 37.43% | 31.39% | 32.84% | **42.24%** |
|  | Recall | **74.80%** | 54.47% | 56.91% | 53.66% | 55.28% |
|  | F1 | 1.45% | 44.37% | 40.46% | 40.74% | **47.89%** |
| **2020/02/09** | Precision | 1.02% | 41.99% | 37.50% | 36.50% | **43.27%** |
|  | Recall | **84.66%** | 40.21% | 55.56% | 50.79% | 47.62% |
|  | F1 | 2.01% | 41.08% | 44.78% | 42.48% | **45.34%** |
| **2020/05/18** | Precision | 11.96% | 66.80% | 70.97% | **73.14%** | 71.70% |
|  | Recall | **89.86%** | 72.19% | 70.97% | 72.28% | 72.57% |
|  | F1 | 21.11% | 69.39% | 70.97% | **72.70%** | 72.13% |
| **AVERAGE** | Precision | 6.39% | 54.79% | 55.88% | 57.80% | **59.54%** |
|  | Recall | **86.55%** | 64.45% | 67.52% | 66.07% | 66.13% |
|  | F1 | 11.55% | 58.79% | 60.46% | 61.12% | **62.56%** |

In the evaluation of different settings of the proposed deep learning network, most results trained with all images get better performances than those with a single image, which indicates that more samples will result in a better optimization of network. The results trained with weighting process shows a slight improvement than those without weighting, which indicates that the weighting has the ability to deal with the imbalance of vehicle and non-vehicle samples. If the model is fine-tuned with the specific image after trained with all images, the results will get obvious improvements for the two images during lockdown. This may be because the fine-tune will make the model more effective in detecting vehicles in the specific image, which is more suitable when the vehicles on the road are rare during lockdown. Therefore, in the following contents, we choose the results with all images for the images acquired on 28 November, 2018, 19 October, 2019, and 18 May, 2020, while select the results after fine-tune for the images on 30 January, 2020, and 9 February, 2020.

*4.2. Experimental Results*

In the experiment, we implemented the proposed vehicle detection algorithm on the whole study site. The vehicles detected are labeled with yellow rectangles. In Figure 7, we show 2 typical regions in these two multi-temporal remote sensing images. Figure 7 (a) - (e) shows Hangkong Road flyover near Wushang Plaza, which is one of the most popular shopping centers in Wuhan. It illustrates that during Wuhan lockdown, the number of vehicles on these roads takes a rapid decrease, and after lockdown lift, the vehicle count increase approaching the traffic situation before lockdown. Furthermore, many detected objects in the image during lockdown belong to false alarms visually, which will lead to low accuracies for these 2 images. Figure 7 (f) - (j) shows Zhuyeshan Road flyover on the 2nd ring road of Wuhan, where there are always heavy traffic jams in weekdays. The decrease of traffic density after the implementation of Wuhan lockdown, and the increase after lockdown lift can be easily observed. It is worth noting that there are lots of miss detections in the heavy traffic jam of Figure 7 (f), since the basic assumption of the proposed method is to distinguish vehicles contrast to road background, which is not satisfied in a traffic jam.

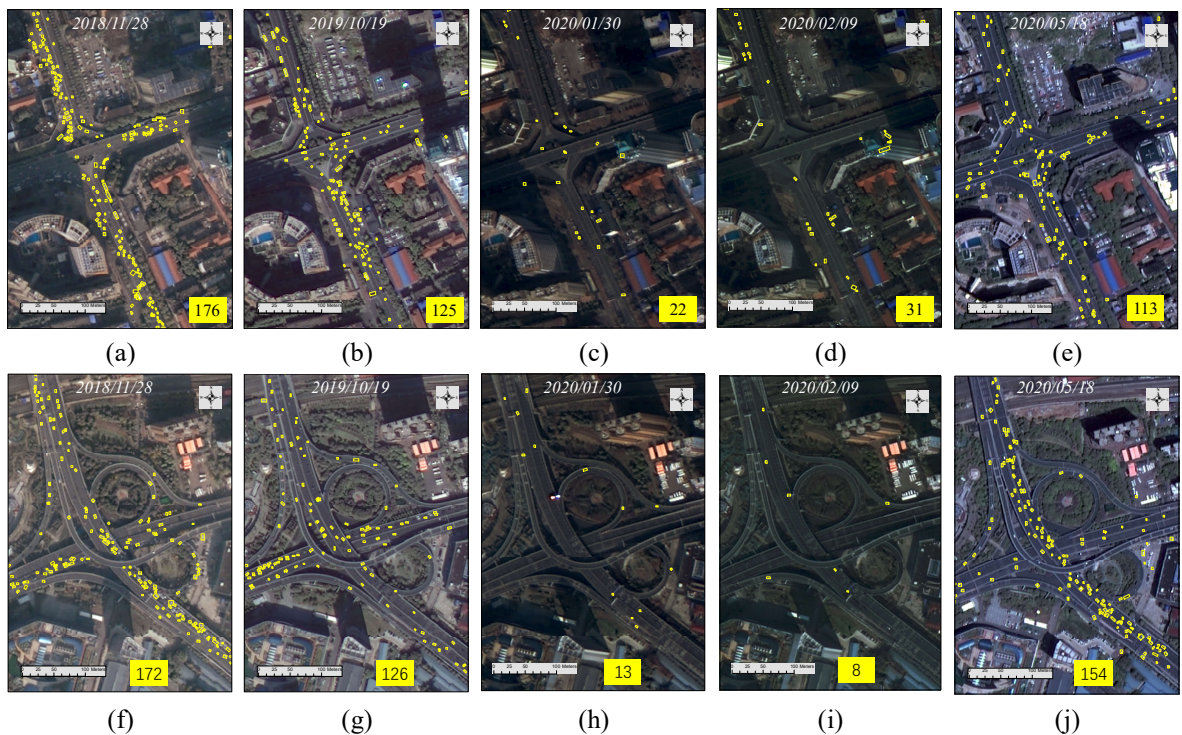

Figure 7. Vehicle detection results of Hangkong Road flyover acquired on (a) 28 November, 2018, (b) 19 October, 2019, (c) 30 January, 2020, (d) 9 February, 2020, and (e) 18 May, 2020; Vehicle detection results of Zhuyeshan Road flyover acquired on (f) 28 November, 2018, (g) 19 October, 2019, (h) 30 January, 2020, (i) 9 February, 2020, and (j) 18 May, 2020. The yellow rectangles indicate the detected vehicles, and the numbers in the bottom right corner indicate the vehicle counts in the images.

The statistics of vehicle numbers and the corresponding dropping percentages are shown in Table 5. The vehicle numbers extracted from image on 28 November, 2018 are regarded as the basis to calculate the dropping percentage. It can be seen that after shadow removal, only the vehicle counts on the last images show obvious reduction. The reason is that the solar zeniths of the other four images are similar and large, while only that of the last image was acquired in summer with a small solar zenith, as shown in Table 1. As a result, the shadow coverages on the first 4 images are also similar and larger than that on the last image. Therefore, shadow removal is meaningful for a fair comparison among these 5 images.

**Table 5 Statistics of counts and dropping percentage of all vehicles and vehicles on main roads before or after shadow removal.**

|  |  | 2018/11/28 | 2019/10/19 | 2020/01/30 | 2020/02/09 | 2020/05/18 |
|---|---|---|---|---|---|---|
| **Before Shadow Removal** | All Vehicle | 52850 | 49391 | 7643 | 10541 | 85679 |
|  | Vehicle On Main Road | 23456 | 18749 | 1661 | 1831 | 25596 |
|  | *Dropping Percentage (All)* |  | *-7%* | *-86%* | *-80%* | *+62%* |
|  | *Dropping Percentage (Main Road)* |  | *-20%* | *-93%* | *-92%* | *+9%* |
| **After Shadow Removal** | All Vehicle | 51284 | 47233 | 7426 | 10118 | 75172 |
|  | Vehicle On Main Road | 22630 | 17759 | 1602 | 1825 | 22862 |
|  | *Dropping Percentage (All)* |  | *-8%* | *-86%* | *-80%* | *+47%* |
|  | *Dropping Percentage (Main Road)* |  | *-22%* | *-93%* | *-92%* | *+1%* |

The changing tendency of vehicle counts is also shown in Figure 8. It can be observed that there is a decrease of traffic density, especially on main roads, on 19 October, 2019, which is due to the weekend impact. After the implementation of Wuhan lockdown, the vehicle numbers dropped with the percentages of higher than 80%, and even higher than 90% on main roads. Considering there were lots of vehicles parked at the roadside during lockdown, which is forbidden in normal days, the dropping percentages for vehicles on main roads may be more accurate to represent the true situation. After lifting Wuhan lockdown, the vehicle number shows an obvious increase to normal days. It is a little strange that the vehicle number is higher than that before lockdown. By observing the detection results in detail, we find that even though many vehicles drive out from parking lots and run on the road, there were still lots of parking vehicles at the roadside on 18 May, 2020. If we look at the dropping percentage on main road, the only 1% increase belongs to the normal variation.

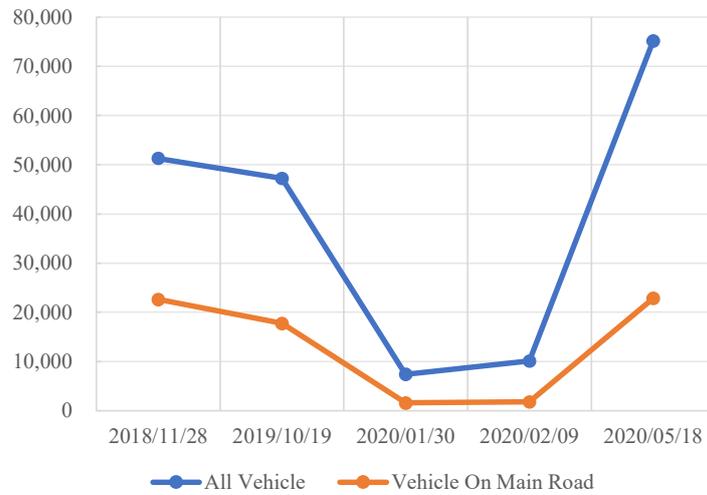

Figure 8. Number changes of all vehicles and vehicles on main roads after shadow removal.

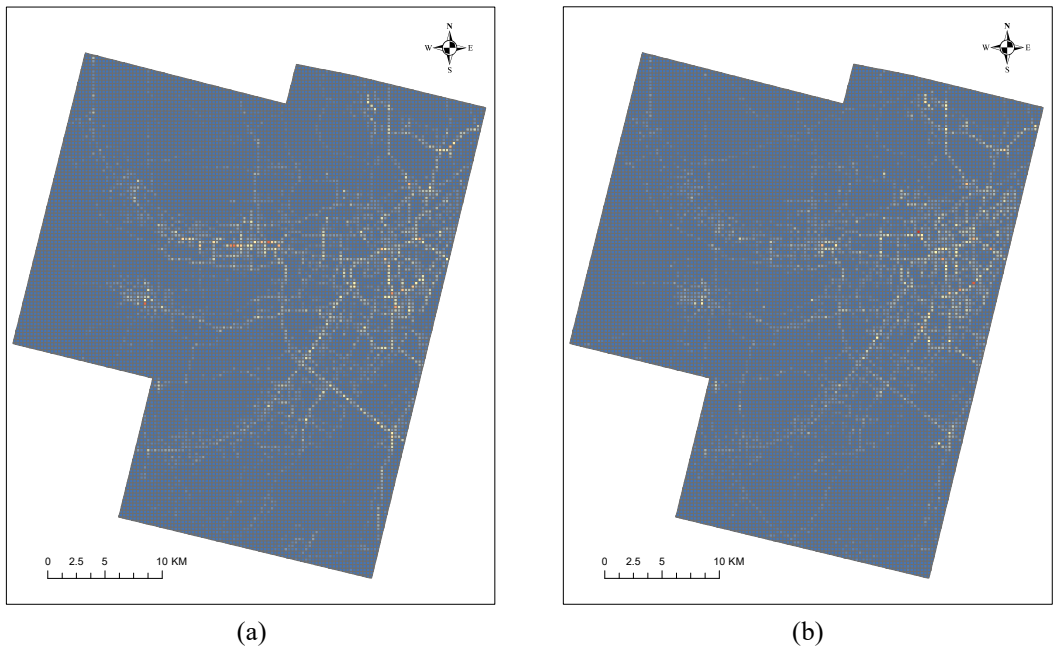

(a)          (b)

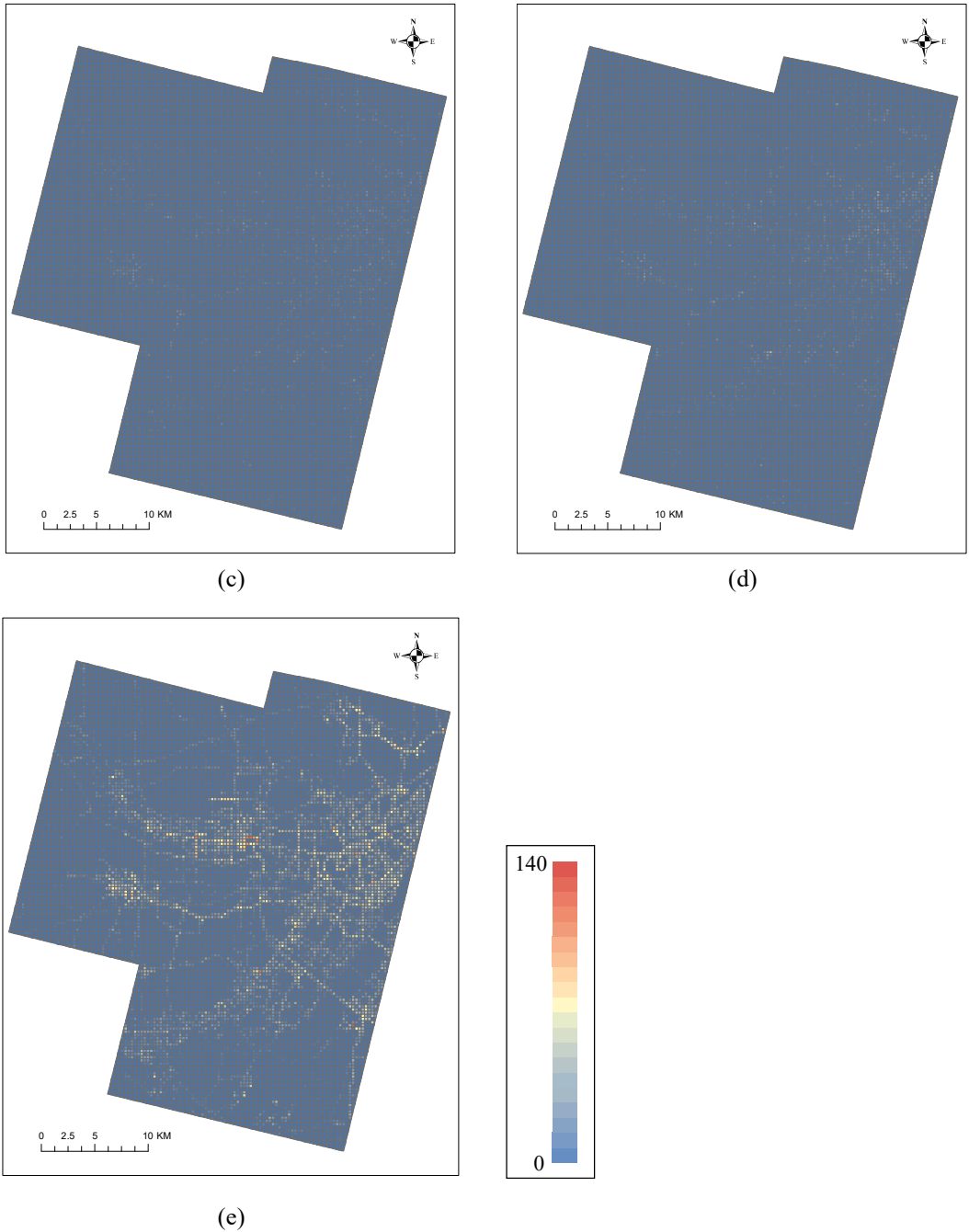

(e)

Figure 9. Statistics of vehicles in $300m \times 300m$ blocks for the images acquired on (a) 28 November, 2018, (b) 19 October, 2019, (c) 30 January, 2020, (d) 9 February, 2020, and (e) 18 May, 2020.

In order to make a global evaluation, we make a statistics of vehicles in $300m \times 300m$ blocks throughout the whole study site after shadow removal. The figures for the statistics before, during and after Wuhan lockdown are shown in Figure 9. In Figure 9 (a), (b) and (e), the distributions of traffic density show the structure and road network of Wuhan city. However, in Figure 9 (c) and (d) during Wuhan lockdown, the vehicle numbers on the road rapidly decrease throughout the whole city. Considering all the public transports had been stopped in the beginning of city lockdown, Figure 9 (a) – (d) provide the evidence that the whole city paused to limit intracity population transmission to mitigate the growth of epidemic. In Figure 9 (e), it can be observed that the transportation resumed normal work in a certain degree after lockdown lift in April.

# 5. Discussion

## 5.1. Accuracy Analysis

Even though we have tried our best to detect vehicles accurately, considering the resolution of remote sensing image, there are still many factors leading to false alarms and omission errors. The main effect factors are summarized as follows:

(1) The resolution of remote sensing image is the main factor affecting accuracy. With the resolution of 0.9m, there is no enough shape information to identify vehicles. The feasible way is to find out vehicles contrast to the road and refine the results considering their neighbor information. However, some interferences, such as road marks and building edges, show quite similar shapes with vehicles, thus it is difficult to distinguish in such a resolution.

(2) During Wuhan lockdown, lots of vehicles were parked at the roadside, which is not allowed in normal days. It is hard to distinguish parking vehicles and running vehicles in remote sensing images. These detected vehicles will lead to the under-estimation of traffic density reduction during lockdown.

(3) In normal days, there will be traffic jams or vehicle parking before traffic light. Since the basic assumption of the proposed method is to find out vehicles contrast to background, the vehicles parked with small spacing in such situations are hard to be extracted. Considering traffic jams and vehicle parking before traffic light didn't exist in the period of lockdown, the reduction of transportation density will also be under-estimated due to this cause.

(4) Since the observation view of these five images are different as shown in Table 1, more obstructions of high buildings will be observed in the image during lockdown, which will result in the reduction of vehicle counts.

By summarizing all the possible factors, it can be found that the traffic density reduction during lockdown is highly possible to be under-estimated. Since the main roads, including high-level roads and ring roads, will have wider roadways, fewer interferences and cannot park vehicles, the changes of traffic density on main roads with the decrease of higher than 90% are more meaningful and representative to study the real situation under transportation ban in Wuhan.

*5.2. Limitations*

It is better to utilize image series with a short interval to evaluate the traffic tendencies during the whole progress of Wuhan lockdown. However, for most remote sensing satellite sensors with the resolution higher than 1m, it is hard to obtain and store time-series images with a very high temporal resolution. Some sensors, such as Planet, can get the temporal images with the frequency of about 7 days, whereas its resolution only reaches 3m, which is hard to distinguish vehicles with a normal size smaller than $1 \times 2$ pixels from other interferences on the roads (Chen et al. 2021). Therefore, by looking at all the available remote sensing images with the resolution of about 1m, we selected GF-2 image series covering the whole period of Wuhan lockdown to study the traffic density tendency.

3 of these 5 images were acquired in weekends, which are comparable for the reduction or increase of traffic density throughout the lockdown period. 2 images were acquired during the beginning of Wuhan lockdown, which can reflect the real traffic situation excluding exception. Although one image was acquired on 9 February, 2020, when is a weekday, it has no influence on the traffic situation due to the strict lockdown policy.

In order to better extract vehicles from the image, we utilize morphology filter to find vehicle candidates and refine the results with deep learning network. The basic assumption is that the vehicles should be different from the road background. However, in the resolution of approximately 1m, some vehicles may be connected to other vehicles, roadsides and other interferences in the view of images. These vehicles will be falsely filtered out due to the shape filter rules. Some interferences, such as road marking paints, is hard to be distinguished with the true vehicles in such a resolution. The accuracy of vehicle extraction will be more seriously affected in the images acquired during lockdown, since fewer vehicles run on the road whereas the interferences still exists. How to improve the accuracy of traffic density estimation in the remote sensing imagery with a normal spatial resolution is the focus of our future work.

## *5.3. Correlation and Difference with Previous Works*

In our previous work (Wu et al. 2021a), we have collected multitemporal high-resolution remote sensing images before and after lockdown implementation covering six cities around the world, including Wuhan, Milan, Madrid, Paris, New York, and London. The traffic density changes caused by lockdown were evaluated and regressed with lockdown stringency to quantify the impact of city lockdown. Since the previous work and this manuscript belong to a series of research, they have some similarities, such as vehicle detection model and the procedure. However, these two works also have different focus and contributions as follows:

1) The previous work focuses on the comparison between different cities around the world with two remote sensing images, while this works focuses on the temporal variations before, during and after Wuhan lockdown with a time-series images. In this work, two images were acquired before city lockdown, two images were acquired during lockdown, and one image was collected after lockdown lift. The reduce of traffic density shows the impact of lockdown policy on limiting intracity population transmission, and the recovery after lockdown lift indicates the life in Wuhan turn into normal in May.
2) These two works are based on different high-resolution remote sensing satellites. The previous work collected Pleiades and Worldview with the resolution of about $0.5m$, while this work used GF-2 images with the resolution of about $1m$. Since the resolution of GF-2 imagery is nearly 2-time lower than Pleiades and Worldview, there are much fewer spatial details of vehicle in this image. Therefore, in this work, we improved the network structure to fit the resolution.

3) In the previous work, in order to fit similar areas of other study cities, we only selected study region with the area of $171.00 km^2$ inside the second ring road of Wuhan. In this work, we choose the maximum common region of the five available multi-temporal images, which contain the area of $1273.61 km^2$, 7.5 times larger than that in the previous work. Therefore, the coverage of Wuhan city in this work is much larger and complete.

## 6. Conclusion

In order to reduce the spread of COVID-19, government enacted a strict lockdown policy in Wuhan from 23 January, 2020, including a transmission control for all inbound /outbound transportation, and a transportation ban policy inside the city (Chinazzi et al. 2020; Kraemer et al. 2020; Tian et al. 2020). Lots of researches have indicated that human mobility control between cities and provinces have slowed the epidemic progression obviously (Jia et al. 2020). Whereas, few studies focus on the transportation ban inside the city, which may be because that the measurement of transportation ban is hard to implemented. Therefore, we collected 5 high-resolution remote sensing image datasets acquired before, during and after Wuhan lockdown by a Chinese satellite GF-2, to analyze the traffic density variation caused by the implementation and lift of transportation ban inside the city.

Due to the resolution of approximately 1m, the traditional object detection methods are unable to detect vehicles without detailed shape information. We utilize a vehicle detection method combining candidate extraction by morphology filter, and vehicle identification by a multi-branch CNN model considering their road background. The experiment and accuracy evaluation show that our method obtained satisfactory performances in these images considering the spatial resolution.

With the experimental results, we find that the traffic density in Wuhan dropped with a percentage higher than 80% after the implementation of lockdown policy. Considering the interferences, the dropping percentage of main roads should be more reasonable with the value higher than 90%. After the lockdown was lifted, the traffic density returned to a normal rate on main roads. Whereas, there are still some vehicles parking at the roadside, which lead to higher vehicle counts on the residential roads. Considering the public transports have been shut down in the period of Wuhan lockdown, the significant reduction of vehicle counts on the roads indicates that the lockdown policy in Wuhan show effectiveness in controlling intracity human transmission. The recovery of traffic density after lockdown lift also show the city returned to normal little by little.

## Acknowledgements

This work was supported in part by the National Natural Science Foundation of China under Grant 61971317, 41801285, and 61822113. We acknowledge Hubei Data and Application Center of High-Resolution Earth Observation System for providing remote sensing images.

# Reference

Audebert, N., Saux, B.L., & Lefevre, S. (2017). Segment-before-Detect: Vehicle Detection and Classification through Semantic Segmentation of Aerial Images. *Remote Sensing, 9*, 368

Axelsson, A., Lindberg, E., Reese, H., & Olsson, H. (2021). Tree species classification using Sentinel-2 imagery and Bayesian inference. *International Journal of Applied Earth Observation and Geoinformation, 100*, 102318

Chen, S., Yang, J., Yang, W., Wang, C., & Bärnighausen, T. (2020). COVID-19 control in China during mass population movements at New Year. *The Lancet, 395*, 764-766

Chen, Y., Qin, R., Zhang, G., & Albanwan, H. (2021). Spatial Temporal Analysis of Traffic Patterns during the COVID-19 Epidemic by Vehicle Detection Using Planet Remote-Sensing Satellite Images. *Remote Sensing, 13*, 208

Chen, Z., Wang, C., Wen, C., Teng, X., Chen, Y., Guan, H., Luo, H., Cao, L., & Li, J. (2016). Vehicle Detection in High-Resolution Aerial Images via Sparse Representation and Superpixels. *IEEE Transactions on Geoscience and Remote Sensing, 54*, 103-116

China Central Television (2019). Chinese News. In

China News (2020). In

Chinazzi, M., Davis, J.T., Ajelli, M., Gioannini, C., Litvinova, M., Merler, S., Pastore y Piontti, A., Rossi, L., Sun, K., Viboud, C., Xiong, X., Yu, H., Halloran, M.E., Longini, I.M., & Vespignani, A. (2020). The effect of travel restrictions on the spread of the 2019 novel coronavirus (2019-nCoV) outbreak. *Science*, 2020.2002.2009.20021261

Eikvil, L., Aurdal, L., & Koren, H. (2009). Classification-based vehicle detection in high-resolution satellite images. *ISPRS Journal of Photogrammetry and Remote Sensing, 64*, 65-72

Gao, J., & O'Neill, B.C. (2020). Mapping global urban land for the 21st century with data-driven simulations and Shared Socioeconomic Pathways. *Nature Communications, 11*, 2302

Gao, S., Rao, J., Kang, Y., Liang, Y., Kruse, J., Dopfer, D., Sethi, A.K., Mandujano Reyes, J.F., Yandell, B.S., & Patz, J.A. (2020). Association of Mobile Phone Location Data Indications of Travel and Stay-at-Home Mandates With COVID-19 Infection Rates in the US. *JAMA Network Open, 3*, e2020485-e2020485

Gotmare, A., Keskar, N.S., Xiong, C., & Socher, R. (2019). A Closer Look at Deep Learning Heuristics: Learning rate restarts, Warmup and Distillation. In, *international conference on learning representations*

Gray, A., Krolikowski, M., Fretwell, P., Convey, P., Peck, L.S., Mendelova, M., Smith, A.G., & Davey, M.P. (2020). Remote sensing reveals Antarctic green snow algae as important terrestrial carbon sink. *Nature Communications, 11*, 2527

Huang, X., Cao, Y., & Li, J. (2020). An automatic change detection method for monitoring newly constructed building areas using time-series multi-view high-resolution optical satellite images. *Remote Sensing of Environment, 244*, 111802

Hubei municipal headquarters for the COVID-19 epidemic prevention and control (2020). Announcement. In

Ji, H., Gao, Z., Mei, T., & Ramesh, B. (2020). Vehicle Detection in Remote Sensing Images Leveraging on Simultaneous Super-Resolution. *IEEE Geoscience and Remote Sensing Letters, 17*, 676-680

Jia, J.S., Lu, X., Yuan, Y., Xu, G., Jia, J., & Christakis, N.A. (2020). Population flow drives spatio-temporal distribution of COVID-19 in China. *Nature*

Jin, S., Yang, L., Danielson, P., Homer, C.G., Fry, J., & Xian, G. (2013). A comprehensive change detection method for updating the National Land Cover Database to circa 2011. *Remote Sensing of Environment, 132*, 159-175

Kingma, D.P., & Ba, J. (2014). Adam: A method for stochastic optimization. In, *arXiv e-prints* (p. arXiv:1412.6980)

Kraemer, M.U.G., Yang, C.-H., Gutierrez, B., Wu, C.-H., Klein, B., Pigott, D.M., du Plessis, L., Faria, N.R., Li, R., Hanage, W.P., Brownstein, J.S., Layan, M., Vespignani, A., Tian, H., Dye, C., Cauchemez, S., Pybus, O., & Scarpino, S.V. (2020). The effect of human mobility and control measures on the COVID-19 epidemic in China. *Science*, 2020.2003.2002.20026708

Leitloff, J., Hinz, S., & Stilla, U. (2010). Vehicle Detection in Very High Resolution Satellite

Images of City Areas. *IEEE Transactions on Geoscience and Remote Sensing, 48*, 2795-2806

Li, Q., Mou, L., Xu, Q., Zhang, Y., & Zhu, X.X. (2019). R3-Net: A Deep Network for Multioriented Vehicle Detection in Aerial Images and Videos. *IEEE Transactions on Geoscience and Remote Sensing, 57*, 5028-5042

Liu, Q., Sha, D., Liu, W., Houser, P., Zhang, L., Hou, R., Lan, H., Flynn, C., Lu, M., Hu, T., & Yang, C. (2020). Spatiotemporal Patterns of COVID-19 Impact on Human Activities and Environment in Mainland China Using Nighttime Light and Air Quality Data. *Remote Sensing, 12*, 1576

Liu, T., Yang, L., & Lunga, D. (2021). Change detection using deep learning approach with object-based image analysis. *Remote Sensing of Environment, 256*, 112308

Mardian, J., Berg, A., & Daneshfar, B. (2021). Evaluating the temporal accuracy of grassland to cropland change detection using multitemporal image analysis. *Remote Sensing of Environment, 255*, 112292

OpenStreetMap Wiki (2020). Key:highway. In

Reichstein, M., Camps-Valls, G., Stevens, B., Jung, M., Denzler, J., Carvalhais, N., & Prabhat (2019). Deep learning and process understanding for data-driven Earth system science. *Nature, 566*, 195-204

Su, H., Yao, W., Wu, Z., Zheng, P., & Du, Q. (2021). Kernel low-rank representation with elastic net for China coastal wetland land cover classification using GF-5 hyperspectral imagery. *ISPRS Journal of Photogrammetry and Remote Sensing, 171*, 238-252

Tang, T., Zhou, S., Deng, Z., Lei, L., & Zou, H. (2017a). Arbitrary-Oriented Vehicle Detection in Aerial Imagery with Single Convolutional Neural Networks. *Remote Sensing, 9*, 1170

Tang, T., Zhou, S., Deng, Z., Zou, H., & Lei, L. (2017b). Vehicle Detection in Aerial Images Based on Region Convolutional Neural Networks and Hard Negative Example Mining. *Sensors, 17*, 336

Tanveer, H., Balz, T., Cigna, F., & Tapete, D. (2020). Monitoring 2011–2020 Traffic Patterns in Wuhan (China) with COSMO-SkyMed SAR, Amidst the 7th CISM Military World Games and COVID-19 Outbreak. *Remote Sensing, 12*, 1636

Tao, C., Mi, L., Li, Y., Qi, J., Xiao, Y., & Zhang, J. (2019). Scene Context-Driven Vehicle Detection in High-Resolution Aerial Images. *IEEE Transactions on Geoscience and Remote Sensing, 57*, 7339-7351

Taubert, F., Fischer, R., Groeneveld, J., Lehmann, S., Müller, M.S., Rödig, E., Wiegand, T., & Huth, A. (2018). Global patterns of tropical forest fragmentation. *Nature, 554*, 519-522

Tian, H., Liu, Y., Li, Y., Wu, C.-H., Chen, B., Kraemer, M.U.G., Li, B., Cai, J., Xu, B., Yang, Q., Wang, B., Yang, P., Cui, Y., Song, Y., Zheng, P., Wang, Q., Bjornstad, O.N., Yang, R., Grenfell, B.T., Pybus, O.G., & Dye, C. (2020). An investigation of transmission control measures during the first 50 days of the COVID-19 epidemic in China. *Science, 368*, 638-642

Tsai, V.J.D. (2006). A comparative study on shadow compensation of color aerial images in invariant color models. *IEEE Transactions on Geoscience and Remote Sensing, 44*, 1661-1671

Wu, C., Zhu, S., Yang, J., Hu, M., Du, B., Zhang, L., Zhang, L., Han, C., & Lan, M. (2021a). Traffic Density Reduction Caused by City Lockdowns Across the World During the COVID-19 Epidemic: From the View of High-Resolution Remote Sensing Imagery. *IEEE Journal of Selected Topics in Applied Earth Observations and Remote Sensing, 14*, 5180-5193

Wu, C., Zhu, S., Yang, J., Hu, M., Du, B., Zhang, L., Zhang, L., Han, C., & Lan, M.J.a.e.-p. (2021b). Transportation Density Reduction Caused by City Lockdowns Across the World during the COVID-19 Epidemic: From the View of High-resolution Remote Sensing Imagery. In   (p. arXiv:2103.01717)

Wuhan municipal headquarters for the COVID-19 epidemic prevention and control (2020a). Announcement No.11 and No. 12. In

Wuhan municipal headquarters for the COVID-19 epidemic prevention and control (2020b). Announcement No. 1. In

Wuhan municipal headquarters for the COVID-19 epidemic prevention and control (2020c). Announcement No. 9. In

Xu, H., Xu, G., Wen, X., Hu, X., & Wang, Y. (2021). Lockdown effects on total suspended solids concentrations in the Lower Min River (China) during COVID-19 using time-series remote sensing images. *International Journal of Applied Earth Observation and Geoinformation, 98*,

Yang, L., Jin, S., Danielson, P., Homer, C.G., Gass, L., Bender, S., Case, A., Costello, C., Dewitz, J., & Fry, J. (2018). A new generation of the United States National Land Cover Database: Requirements, research priorities, design, and implementation strategies. *ISPRS Journal of Photogrammetry and Remote Sensing, 146*, 108-123